\documentclass{article}

\usepackage[preprint]{neurips_2019}

\usepackage[utf8]{inputenc} 
\usepackage[T1]{fontenc}    
\usepackage{hyperref}       
\usepackage{url}            
\usepackage{booktabs}       
\usepackage{amsfonts}       
\usepackage{nicefrac}       
\usepackage{microtype}      
\usepackage{graphicx}
\usepackage{subfigure}
\usepackage{amsmath}
\usepackage{amsthm}
\usepackage{enumitem}
\usepackage{color}
\usepackage{appendix}
\usepackage{caption}

\DeclareMathOperator*{\argmin}{arg\,min}

\newtheorem{theorem}{Theorem}

\title{RL4health: Crowdsourcing Reinforcement Learning for Knee Replacement Pathway Optimization}

\author{ Hao Lu\\
Department of Operations Research and Financial Engineering\\
Princeton University\\
Princeton, NJ, 08544\\
\texttt{haolu@princeton.edu}\\
\AND
Mengdi Wang\\
Department of Operations Research and Financial Engineering\\
Princeton University\\
Princeton, NJ, 08544\\
\texttt{mengdiw@princeton.edu}\\
}

\begin{document}

\maketitle

\begin{abstract}
Joint replacement is the most common inpatient surgical treatment in the US.  We investigate the clinical pathway optimization for knee replacement, which is a sequential decision process from onset to recovery. Based on episodic claims from previous cases, we view the pathway optimization as an intelligence crowdsourcing problem and learn the optimal decision policy from data by imitating the best expert at every intermediate state.
We develop a reinforcement learning-based pipeline that uses value iteration, state compression and aggregation learning, kernel representation and cross validation to predict the best treatment policy. It also provides forecast of the clinical pathway under the optimized policy. Empirical validation shows that the optimized policy reduces the overall cost by 7 percent and reduces the excessive cost premium by 33 percent.
\end{abstract}

\section{Introduction}
Driven by the explosion of computing power and big data, the past decade has witnessed tremendous success of artificial intelligence (AI)  in many applications, like image recognition \citep{russakovsky2015imagenet}, natural language processing \citep{manning2014stanford}, game AI \citep{mnih2015human}, and speech recognition \citep{hinton2012deep}.  Meanwhile, the healthcare industry, with massive volumes of clinical data together with sophisticated devices and recording systems, stands to benefit from this new era of artificial intelligence. Machine learning technologies have proved effective for analyzing medical data at certain scenarios and providing end-to-end predictions; see for examples \citep{esteva2017dermatologist, shickel2018deep, raghu2017continuous}. 

However, artificial intelligence is way underdeveloped for clinical decision making in complex situations. People face tremendous challenges. Healthcare data are featured for high-dimensionality, heterogeneity, temporal dependency, and irregularity \citep{luo2016big, miotto2017deep, doorhof2018using}. Machine learning methods can produce good predictions at certain scenarios, but their model often lacks generalizability and global optimization.  The healthcare system often involves ineffective services, excessive cost, and uncoordinated cares \citep{olsen2010healthcare}.  Conventional machine learning methods cannot prescribe every clinical decision in a systematic way. There lacks a principled AI-driven approach to optimize the full clinical decision process for complicated medical treatments.

In this paper, we focus on the clinical decision process of knee replacement. During an episode of knee replacement treatment, a patient has to go through a sequence of diagnosis, prescriptions and treatments, which we refer to as a clinical pathway. 
We develop a reinforcement learning approach to optimize the knee replacement pathway based on claims records. Empirical results demonstrate that our solution achieves significant improvement over the baseline.

\paragraph{Knee Replacement}
Knee replacement is a surgical procedure using metal and plastic parts to resurface damaged knees. An episode of knee replacement treatment takes approximately three months. During this process, a patient go through a pathway from onset to surgery, rehabilitation and recovery. According to the Centers for Medicare and Medicaid Services (CMS) \citep{comprehensivecareforjointreplacementmodel}, hip and knee replacements are the most common inpatient surgery in the United States. Around 700,000 knee replacement procedures take place every year \citep{martin2017patient}. The average cost per episode ranges from \$16,500 to \$33,000 across different geographical areas \citep{comprehensivecareforjointreplacementmodel}.  
Patients who receive the knee replacement treatments often have to go through a long recovery period. Knee replacement treatments are costly for the patients and the Medicare, and the quality of treatments vary largely among healthcare providers. 

Beginning on April 1, 2016, CMS started a new payment model, called the Comprehensive Care for Joint Replacement (CJR) model \citep{comprehensivecareforjointreplacementmodel}, to reduce excessive costs and support better care for patients are undergoing elective hip and knee replacement surgeries.
The CJR model will run for five years in 67 geographic areas. 
 According to the CJR, more than 800 hospitals are required to keep the knee replacement cost below \$25,565 for each episode, and they will face a financial penalty for exceeding this threshold \citep{japsen_2016}. 
According to \citep{barnett2019two}, after two years since it was implemented, the CJR model achieved a cost reduction of 3.1\% per episode. 
Motivated by the pressing needs for cost reduction and better care, we aim to optimize the clinical pathway of knee replacements, by leveraging the reinforcement learning  technology to learn the best clinical decisions from data.
\paragraph{Our Approach}
Our starting point is a data set consisting full episodes of claims records from hundreds of successful past treatments. The records are collected from different care providers who are all experts in treating knee replacements. We view the clinical pathway optimization as an {\it intelligence crowdsourcing} problem. Care providers may have different strengths - some of them might make better decisions at certain stages of the process. Our goal is to decompose the treatment process into individual decisions at all possible ``states" and ``imitate" the best expert at every state of the process.

We develop a reinforcement learning-based approach to learn the optimal policy from episodic claims. 
Reinforcement learning is a branch of artificial intelligence \citep{sutton1998introduction}. It is well-suited for the sequential, uncertain and variable nature of clinical decision making. By modeling the knee replacement episode as a Markov decision process, we develop a reinforcement learning pipeline to find the optimal treatment decision with exponentially many states and actions and takes its long-term effect into account. 
Our main contributions are as follows:

\begin{itemize}[topsep=0pt,itemsep=2pt,parsep=0pt,partopsep=0pt, leftmargin = 20pt]

\item We develop a reinforcement learning-based pipeline to analyze episodic clinical claims records and synthesize expertise of multiple physicians into an optimal treatment policy. The pipeline provides forecast of future pathway at any intermediate stage under the optimal policy.
 
\item The pipeline employs a dimension reduction technique to automatically aggregate histories and diagnosis in a way to maximally preserve the quality of the solution, by leveraging the intrinsic spectral properties of the underlying transition probability distributions. It also uses cross validation to find the most robust and generalizable solution even if data is limited, as well as to provide confidence bound of the predicted performance.

\item The optimized policy reduces the average cost per episode by \$1,000, which is approximately 7\% of the full cost. For cost exceeding the repayment threshold, the optimized policy reduces the average excessive premium by \$500, which is approximately 33\% of the excessive cost. 
\end{itemize}

If successfully validated and implemented, the optimized treatment policy is projected to result in 700 millions in savings each year in the US. 
The proposed approach provides a prototype for machine learning-based clinical pathway optimization. We believe it will apply to a broader scope of treatments and can significantly improve the outcome of care, leading to cost reduction and better patient experience.
\vspace{-0.1cm}
\section{Related Work}
\vspace{-0.1cm}

Reinforcement learning (RL) is an area of machine learning for solving sequential decision-making problems from data or experiment \citep{sutton1998introduction}. RL has achieved phenomenal successes in many fields, for examples \citep{mnih2015human, silver2017mastering, li2010contextual, sutton2012dyna, li2017deep}. Compared with conventional machine learning, RL focuses on learning and acting in a dynamic environment, which is powerful to model and recommend sequential decisions to be made dynamically in clinical treatments.

There have been several attempts to apply RL in healthcare scenarios. For sepsis treatment optimization, \citep{raghu2017deep, raghu2017continuous} proposed a continuous-state framework to generate optimal treatment policies with deep reinforcement learning, and \citep{peng2018improving} developed a mixture-of-experts framework by combining neighbor-based algorithms and deep RL. \citep{nemati2016optimal} modeled heparin dosing as a Partially Observed Markov Decision Process (POMDP), and designed a deep RL algorithm to learn individualized heparin dosing policy. \citep{prasad2017reinforcement} used off-policy reinforcement learning to improve the dosing policy in ICU. \citep{parbhoo2017combining} used the kernel-based regression and dynamic programming to improve therapy selection for treating HIV. In these applications, the decision is often the dosage, which is a single-dimensional variable, or the decision is restricted to a small number of choices to reduce the problem's complexity.

For knee replacement treatment, there exist studies where machine learning approaches are used for prediction or classification; see \citep{huber2019predicting, navarro2018machine, cabitza2018machine}. However, these studies did not consider the optimization of clinical decision process.

Our work appears to be the first one to systematically optimize the knee replacement treatment. 
The proposed approach uses ideas including kernel-based reinforcement learning, unsupervised state aggregation learning and dimension reduction. Kernel-based reinforcement learning was proposed in 2002 by \citep{ormoneit2002kernel}. A kernel function is needed to measure the similarity between states, based on which one can approximately solve high-dimensional Markov decision problems. It is shown that kernel-based RL produces a convergent solution that also achieves statistically consistency. Their model has been studied and extended in a number of works \citep{jong2006kernel, kveton2013structured, barreto2016practical}. 
A critical component of our approach is to learn the right kernel automatically from data.
We design the kernel function by using unsupervised spectral state compression and aggregation learning, which were proposed by \citep{zhang2018spectral, duan2018state}. They provide a method to estimate the optimal partition of state space from trajectoric data to maximally preserve the transition dynamics in the data set. In this way, we are able to remedy the curse of dimensionality of high-dimensional RL by using features that are automatically learned from data.

Our approach provides a novel end-to-end pipeline for clinical pathway optimization from clinical data. It allows one to learn from records of a group of medical experts to achieve intelligence crowdsourcing by using statistical dimension reduction and reinforcement learning. The outcome is a full decision policy with robustness guarantee, together with a predictive model to forecast a patient's clinical pathway beginning from any intermediate state of the process.

\section{Clinical Pathway Optimization Model}

In this section, we describe the sequential decision model of the knee replacement treatment.

\textbf{Knee Replacement Treatment as a Markov Decision Process (MDP)}. ~We model the clinical decision making process of knee replacement as an infinite-horizon Markov decision process (MDP) with an absorbing terminal state. MDP models a dynamic state-transition process, where the state $s_t$ at time $t$ evolves to a future state $s_{t+1}$ according to a transition law under the intervention of a decision maker.  An instance of MDP can be described by a tuple $\mathcal{M} = (\mathcal{S}, \mathcal{A}, P, C)$, where $\mathcal{S}$ is set of all possible states, $\mathcal{A}$ is set of all possible actions, $P$ is a law of transition which is not explicitly known to the decision maker, and $C:\mathcal{S}\times\mathcal{A} \rightarrow \mathbb{R}$ is a state-transition cost function. An episode in an MDP is a sequence of states, actions and costs which ends at a terminal state. In the case of knee replacement, each episode is a full claims record for a patient's pathway from onset to rehabilitation and recovery. Each claim has one diagnosis and one procedure (also called prescription). We model each claim as a time step. The knee replacement process may last for an indefinitely number of time steps. We model ``recovery" as an absorbing terminal state, at which no future transition or cost will be generated. 
In a given episode, ideally speaking, a state $s_t$ is a collection of claims up to the time step $t$, and an action $a_t$ is picked from all possible prescriptions. 
At time $t$, a physician examines the current state $s_t\in\mathcal{S}$ of a patient, chooses an action $a_t\in\mathcal{A}$ according to his/her own expertise and then the patient moves to the next state $s_{t+1}$ according to a probability transition law $P(s_{t+1} \mid s_t,a_t)$. Each claim may generate a noisy cost $C(s_t,a_t)$. Figure \ref{illus} shows the model of an infinite-horizon Markov decision process (MDP) in the case of knee replacement treatment.
 
\begin{figure}[t]\label{illus}
\centering
\includegraphics[width= 5in ]{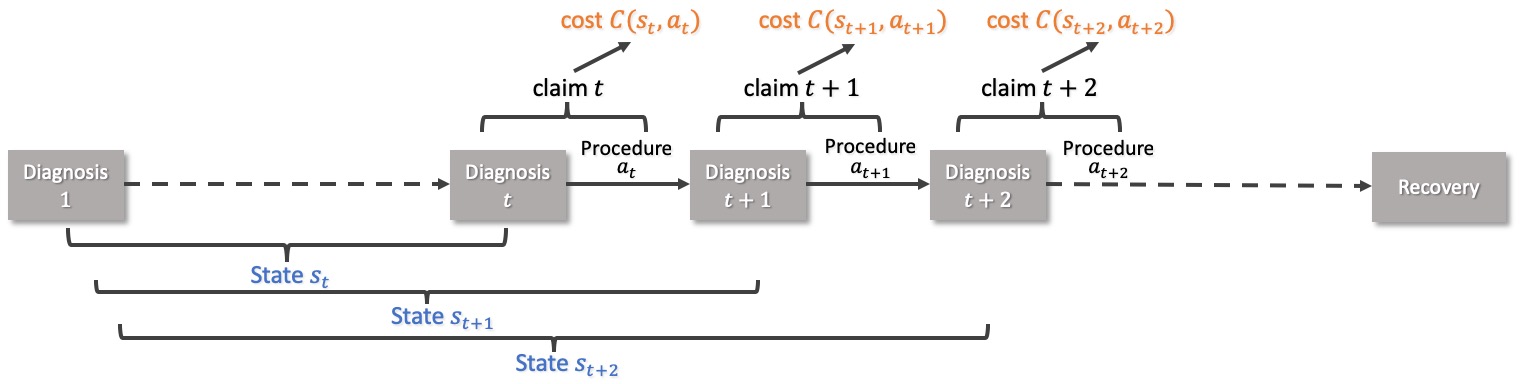}\caption{\footnotesize Knee replacement episode as a Markov decision process (MDP) with absorbing state (recovery).\label{illus}}
\end{figure} 
\textbf{Treatment Policy and Objective}. ~The treatment policy is modeled as a function that maps a state (claims history) to an action (a prescription), which we denote by $\pi: \mathcal{S} \rightarrow \mathcal{A}$. Our goal is to find an optimal policy to minimize the expected cumulated cost per episode, i.e., to solve the optimization problem
\vspace{-0.1cm}
\begin{align*}
\min_{\pi} \mathbb{E}^{\pi}\left[\sum_{t=1}^{\infty} C(s_t,a_t)\right],
\end{align*}

where $\mathbb{E}^{\pi}$ is taken over the entire pathway $(s_1,a_1,s_2,a_2,\ldots)$ on which actions are chosen according to $\pi$.
Solving this ideal optimization problem precisely is intractable because the transition probabilities $\{P(s'|s,a)\}$ are unknown. Therefore, we propose a compression-kernel method to estimate the probability transition law using low-dimensional models (See Section \ref{method} for more details).

\textbf{Value Functions and Bellman Equation}. ~For any state $s\in\mathcal{S}$, we define its value function of a policy $\pi$ as $V^{\pi}(s) = \mathbb{E}^{\pi}[\sum_{t=1}^{\infty} C(s_t,a_t)| s_0 = s]$. From the optimal control literature \citep{bertsekas1995dynamic} we know that
there always exists an optimal policy $\pi^*$, whose value vector $V^*(s)\leq V^{\pi}(s)$ for any policy $\pi$ and state $s$, and this optimal value vector $V^*$ satisfies the following {\it Bellman equation}:
\vspace{-0.1cm}
\begin{align}\label{Bellman}
V^*(s) = \min_{a\in \mathcal{A}}\bigg[C(s,a) + \sum_{s'\in\mathcal{S}}P(s'|s,a)\cdot V^*(s')\bigg],~~~~\forall s\in\mathcal{S}.
\vspace{-0.15cm}
\end{align}

Section \ref{method} shows how to estimate the optimal policy by solving the high-dimensional Bellman equation from empirical data\eqref{Bellman}. 
 
\textbf{State Space of Knee Replacement Treatment}. ~Given $138$ different diagnosis categories and around $30$ claims per episode,
the number of possible states $|\mathcal{S}|$ can be  as large as $138^{30} \approx 10^{64}$. The number of possible actions is $597$ according to the data set. Such huge numbers of states and actions mean that the Bellman equation is of huge dimensions, making basic reinforcement learning methods intractable.
Therefore we reformulate the policy optimization problem in the spirit of intelligence crowdsourcing.
We model the state as 
\vspace{-0.15cm}
\begin{center}$s_t =$ (most recent diagnosis up to time $t$, {number of times one has been inpatient up to time $t$}).
\end{center} 
\vspace{-0.15cm}
The reformulated state is still sufficient to capture most of the information needed for making decisions. However, it still yields a large-scale MDP with more than 600 unique states. Further dimension reduction is needed (to be discussed in Section \ref{method}).

\textbf{Intelligence Crowdsourcing and Prescription Policy}. ~In this paper, to reduce the dimension of action space $\mathcal{A}$, we split the physicians randomly into multiple groups $1,\ldots, J$ and make sure that these groups have identical average cost per episode. Suppose the current state is $s_t$, we model the action $a_t \in \mathcal{A} := \{1,\ldots,J\}$ to be: First pick a group of physicians from $1,\ldots,J$, then pick one doctor from the group uniformly at random, finally imitate the prescription of this physician at the current state $s_t$. Thus the resulting state-to-prescription policy is based on the selected doctor's conditional distributions of prescriptions.  
Since we have reduced $\mathcal{A}$ to be a small finite set, the action complexity has been reduced significantly. More importantly, this model can maximally leverage the physicians' expertise from their past experiences.

\section{Method}\label{method}

In this section we present our reinforcement learning-based pipeline with state compression, kernel representation and value iteration.

\subsection{Unsupervised State Compression and Kernel Construction}
To find a robust policy from limited noisy data, we need dimension reduction tools to identify latent features of the knee replacement process and further reduce the model complexity. 


We employ an unsupervised spectral state compression method, which was proposed by \citep{zhang2018spectral, duan2018state}. The goal is to estimate leading features of a state-transition process $(X_1,X_2,\ldots,X_t)$ from finite-length trajectories. It is motivated by a ubiquitous structure of high-dimensional state-transition system, that is the transition kernel often admits a low-rank or nearly low-rank decomposition:
$
P(X' \mid X) \approx \sum_z P(X' \mid Z=z) P(Z=z \mid X),
$where $Z$ can be viewed as a latent ``mode".
The low-rank nature makes it possible to estimate the left and right singular functions of the unknown transition kernel $P$ from data. 

Our first step is to compute the empirical transition frequency matrix $F \in \mathbb{R}^{|\mathcal{S}|\times |\mathcal{S}|}$ from the data set. We order all the claims according to claim id in each episode. Every two consecutive claims gives a state-transition pair $(s,s')$. We compute the entry $F_{ss'}$ to be the frequency for $(s',s)$ to appear in the entire data set. After obtaining $F$, we normalize it into a empirical transition matrix $\widetilde P$, such that each row is nonnegative and sums to 1. Then we apply singular value thresholding to the transition matrix $\widetilde P = U\Sigma V$ and get the right top $k$ singular vectors of $\widetilde P$, where $k$ is a tuning parameter. It gives a matrix $\widehat{V} \in \mathbb{R}^{|\mathcal{S}|\times k}$.  According to \citep{zhang2018spectral}, $\widehat{V}$ are referred to as Markov features that are representative of the leading structures of the unknown dynamics. 



Following \citep{zhang2018spectral, duan2018state}, we compute the optimal partition of state space $\mathcal{S} = \mathcal{S}_1\cup\mathcal{S}_2\cup\ldots\cup\mathcal{S}_k$ such that 
$P(\cdot \mid s) \approx P(\cdot \mid s')$ if $s,s' \in \mathcal{S}_j$ for some $j\in[k].$
In this way, we can aggregate states into clusters and find one best action for all states belonging to the same cluster. 
We compute the partition by solving the optimization problem and using the estimated $\widehat{V}$:
\vspace{-1mm}
\begin{align}\label{partition}
 \min_{\mathcal{S}_1, \dots, \mathcal{S}_k} \min _{v_1,v_2,\ldots,v_k\in \mathbb{R}^k}\sum_{m=1}^k\sum_{s\in\mathcal{S}_m}||(\widehat{V})_{[s, :]} - v_m||^2_2.
\end{align} 
In the experiment, we solve the preceding problem by applying the k-means method with 100 random initializations and choosing the best result.  


Now that we have estimated the Markov features $\widehat{V}$ and the optimal partition by using spectral state compression.  We use the estimated partition from \eqref{partition} and construct the state-wise kernel function $K(\cdot,\cdot)$ by 
\begin{align}\label{kernel}
K(s_1,s_2) =
\begin{cases}
1,& \text{if~} s_1,s_2 \text{~belongs to the same block},  \\
0,& \text{otherwise}. 
\end{cases}
\end{align}
Alternatively, we can define the kernel function by $K(s_1,s_2) = \widehat{V}_{[s_1, :]}^T \widehat{V}_{[s_2, :]},$ which means that we approximate the transition law within the Hilbert space spanned by its principal components.
\vspace{-0.2cm}
\subsection{Kernel-Based Model Estimation}\label{empirical MDP}
\vspace{-0.1cm}
So far we have obtained a kernel function $K(\cdot,\cdot)$  that captures the similarity between any two states. It is a non-negative mapping defined as $K: \mathcal{S}\times \mathcal{S} \rightarrow \mathbb{R}^+$.
Now we can use the kernel function to construct an empirical MDP model $\widehat{\mathcal{M}} = (\mathcal{S}, \mathcal{A}, \widehat{P}, \widehat{C})$ from the data, which is defined as following.  


Let $\mathcal{D} = \{(s_m,a_m,c_m,s'_m)| m = 1,2,\ldots,n\}$ be the set of sample transition quadruples (state, action, cost, next state). Here $n$ is the total number of such samples. Given a kernel function $K(\cdot,\cdot)$, we define the empirical MDP as $\widehat{\mathcal{M}} = (\mathcal{S}, \mathcal{A}, \widehat{P}, \widehat{C})$, where $\mathcal{S}, \mathcal{A}$ are the same state space and action space as in the original MDP $\mathcal{M}$. The state-transition probability matrix $\widehat{P}$ is estimated as
\begin{align}\label{kernel transition}
\widehat{P}(s'|s,a) = \frac{\sum_{\{m| a_m = a\}}K(s,s_m)\cdot K(s',s'_m)}{\big[\sum_{\{m| a_m = a \}}K(s,s_m)\big]\cdot \big[\sum_{\{m| a_m = a \}}K(s',s'_m)\big]}
\end{align}
for any $s, s'\in\mathcal{S}$ and $a\in\mathcal{A}$, as long as $s$ is not the terminal recovery state. If $s$ is the terminal state, we always let $\widehat{P}(s|a,s) =1$ and $\widehat{P}(s'|s,a) =0$ for all $a\in\mathcal{A}$ and $s'\not=s$.
The kernelized cost function $\widehat{C}$ is estimated as 
\begin{align}\label{kernel reward}
\widehat{C}(s,a) =\frac{\sum_{\{m| a_m =  a \}}K(s,s_m)\cdot c_m}{\sum_{\{m| a_m =  a \}}K(s,s_m) }
\end{align}
for any $s \in\mathcal{S}$ that is not the terminal state and $a\in\mathcal{A}$, and we let $\widehat{C}(\hbox{terminal},a) =0$ for all $a$.

%
\vspace{-0.1cm}
\subsection{Computing the Optimal Policy}
\vspace{-0.1cm}
By using the appropriate kernel function, we have computed the empirical transition model \eqref{kernel transition} and empirical cost function \eqref{kernel reward} from claims data. In other words, we have obtained an empirical instance of $\widehat{\mathcal{M}} = (\mathcal{S}, \mathcal{A}, \widehat{P}, \widehat{C})$, which is an approximation to the unknown true MDP model $\mathcal{M}$. 

Finally, we solve this empirical MDP problem $\widehat{\mathcal{M}}$ by using the value iteration method \citep{bertsekas1995dynamic}.
Recall the Bellman equation in \eqref{Bellman}, the value iteration makes the following iterative updates on a value vector $V\in\mathbb{R}^{|\mathcal{S}|} $ until it converges to a fixed point:
\begin{align*}\label{operator}
V(s) \leftarrow \min_{a\in \mathcal{A}}\bigg[\widehat{C}(s,a) + \sum_{s'\in\mathcal{S}} \widehat{P}(s'|s,a)\cdot V(s')\bigg],\ \  \forall s\in \mathcal{S},
\end{align*} 
with the terminal condition $V(\text{terminal})=0$.
The value iteration is known to provide a sequence of value functions converging to the optimal value function $V^*\in \mathbb{R}^{|\mathcal{S}|}$. Then we obtain the optimal policy $\pi^*$: 
\vspace{-1mm}
\begin{align*}
\pi^*(s) = \argmin_{a\in \mathcal{A}}\bigg[\widehat{C}(s,a) + \sum_{s'\in\mathcal{S}} \widehat{P}(s'|s,a)\cdot V^*(s')\bigg].
\end{align*}
The computed $\pi^*$ prescribes a physician group to follow for each state. The final treatment policy is a set of conditional distributions over prescriptions, which are estimated from the data set following $\pi^*$. 


\begin{figure}[t]
\centering
\includegraphics[width= 2.82in ]{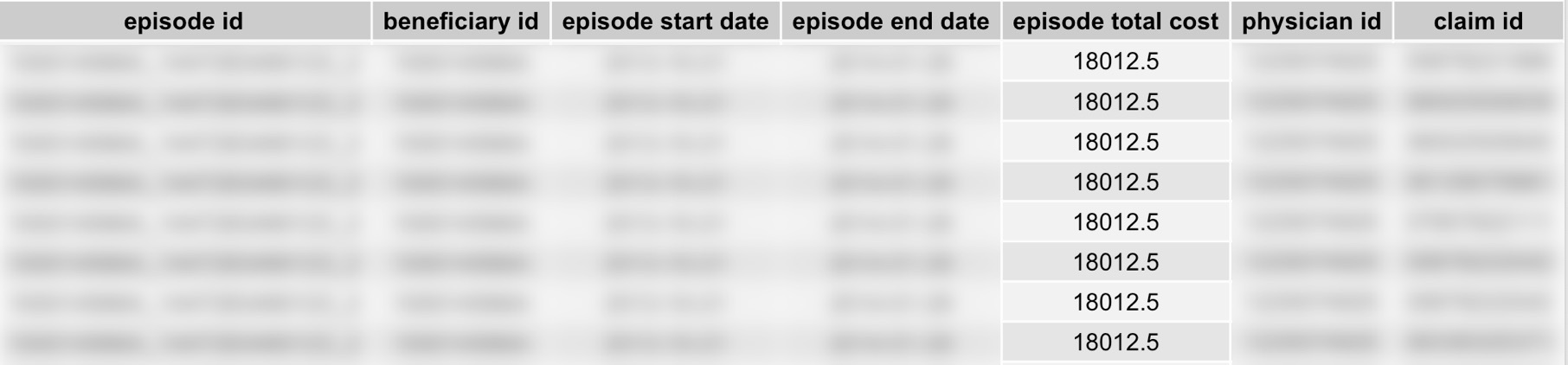}\includegraphics[width= 2.603in ]{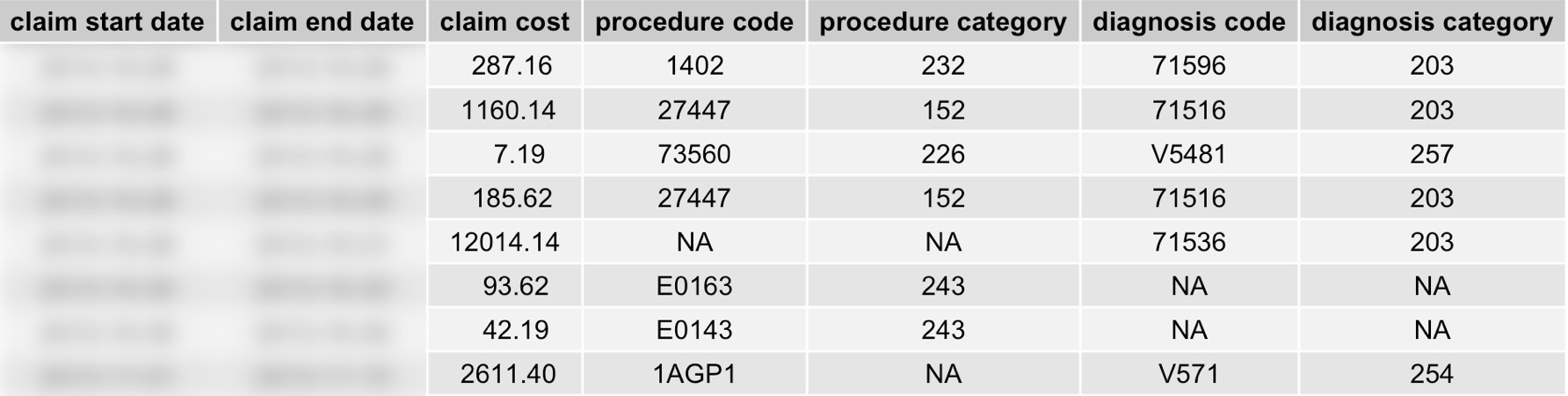}
\setlength{\belowcaptionskip}{-8pt}
\caption{First eight lines of the knee replacement claims data.}
\label{data}
\end{figure}

\vspace{-0.2cm}
\section{Experiments Using Claims Records}\label{experiment}
\vspace{-0.1cm}

In this section, we present our main results. Empirical validation with real knee data shows that the proposed pipeline produces a robust policy and results in significant cost reduction.

We use a data set of knee replacement claims provided by Cedar Gate Technologies. The data set contains claims records from 37 physicians and 205 unique beneficiaries (patients). There are 212 episodes, 5946 claims and 9254 entries. Each episode is a full claims record for a patient's pathway from onset to rehabilitation and recovery. Each episode involves one physician, one beneficiary, a total cost, an episode start date, an episode end date, and a sequence of claims. 
Figure \ref{data} gives a snapshot of several lines in the data set. Each entry contains the following attributes: episode id, beneficiary id, episode start date, episode end date, episode total cost, physician id, claim id, claim start date, claim end date, claim cost, procedure code, procedure category, diagnosis code, and diagnosis category. We have masked the sensitive information to protect the patient's identity. In Appendix \ref{appendix} we will discuss more details about the data set.
\begin{figure*}[t]
\begin{tabular}{cc}
\includegraphics[width= 2.62 in, height = 1.6in ]{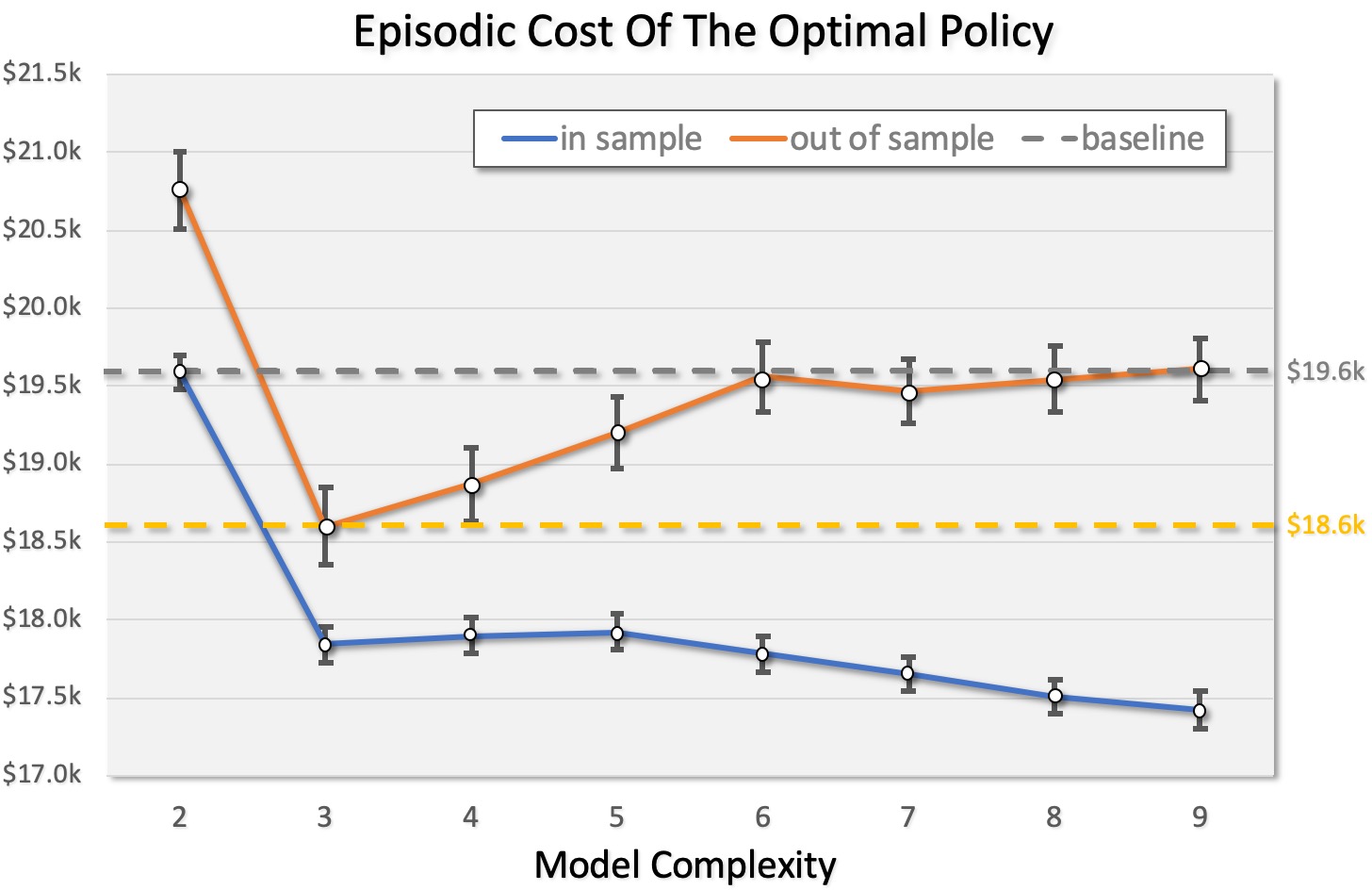} & 
\includegraphics[width= 2.62 in, height = 1.6in ]{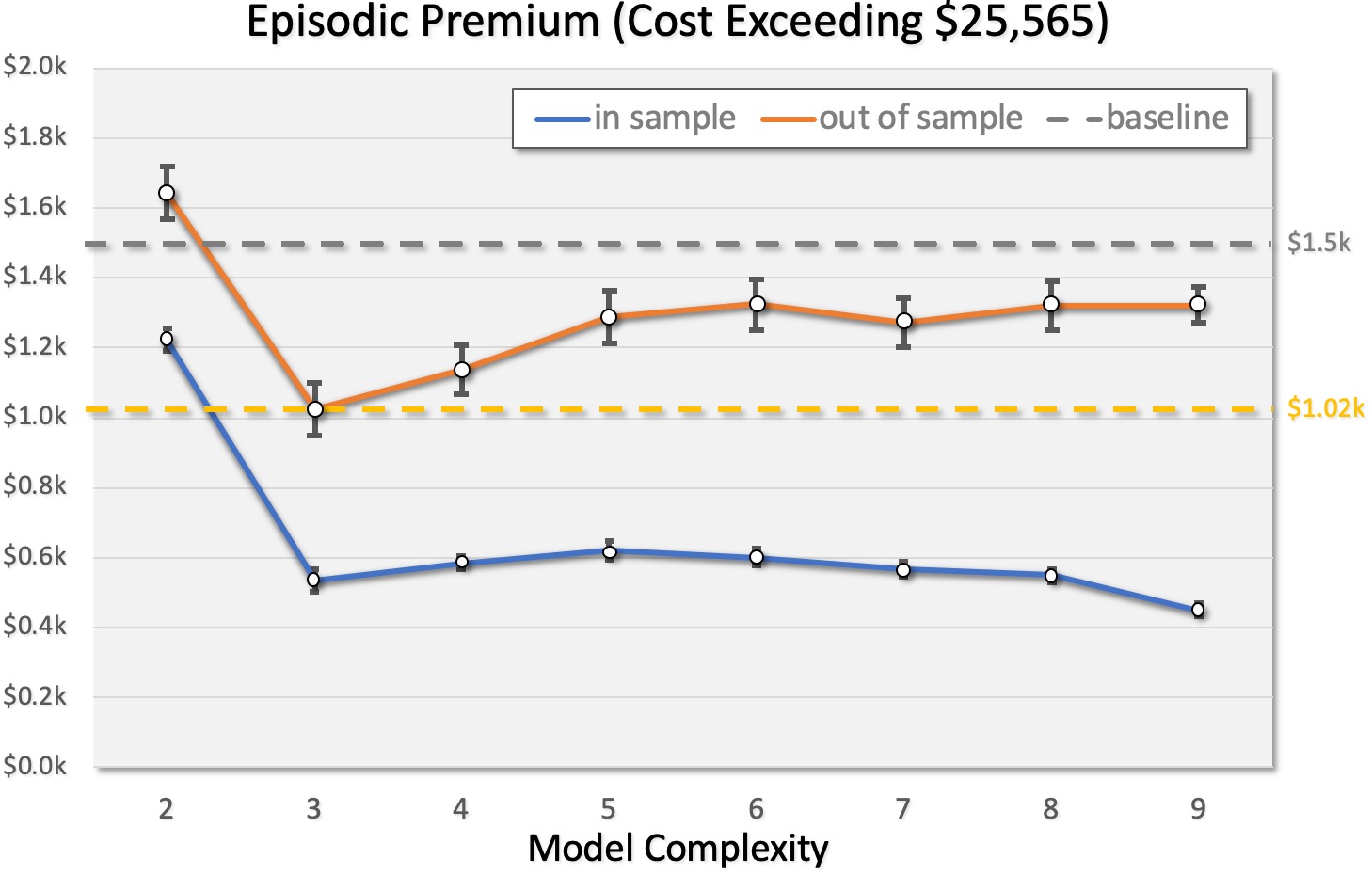}
\\
(a) & (b)\\
\end{tabular}
\setlength{\abovecaptionskip}{5pt}
\setlength{\belowcaptionskip}{-5pt}
\caption{\footnotesize{{\bf In-sample vs out-of-sample performance of the learned optimal policy.} (a): This figure illustrates the in-sample and out-of-sample performances of the optimized policy with varying degrees of model complexity, as well as their 95\% confidence intervals. The baseline (grey) is the average episodic cost in the raw data set. These curves suggest that the model tends to overfit the data as $k>3$. The optimized policy (yellow) is projected to achieve a cost reduction of \$1,000  per episode. (b): This figure illustrates the in-sample and out-of-sample performances of excessive cost premium per episode of the best trained policy. The baseline (grey) is the average episodic premium in the raw data set. The yellow line is the best out-of-sample performance, which is achieved when the model complexity is $k=3$. In this case, the optimized policy is projected to reduce the excessive premium per episode by approximately \$500 or 33\% on average. }}
\label{normal}
\end{figure*}

\begin{figure*}
\centering
\centerline{\includegraphics[width= 2.6in ]{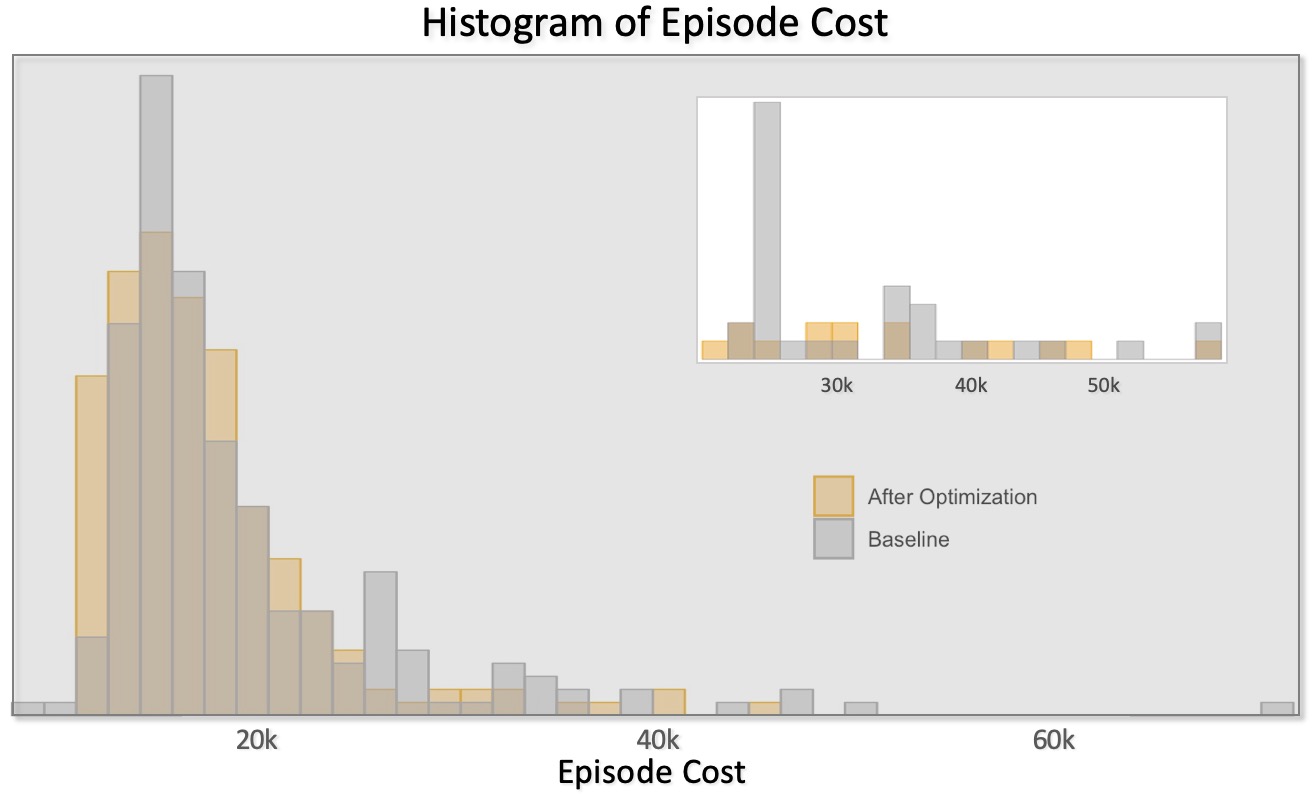}}
\setlength{\belowcaptionskip}{-10pt}
\caption{\footnotesize{{\bf Histogram of costs per episode.} This plot gives the distribution histogram of cost per episode in the raw data (grey) and the simulated cost histogram if the optimized policy is used (yellow). The baseline cost per episode is on average \$19,559 with standard deviation \$8,101, while the improved episode cost after optimization is on average \$17,781 with standard deviation \$5,750. In the upper right corner, we zoom in the tail distribution and examine the costs above \$25,565. The excessive cost premium is \$1,500 in the raw data, which reduces to \$566 after optimization. }}
\label{episode cost}
\end{figure*}
\vspace{-0.2cm}
\subsection{Projected Performances and Cross Validation} \label{project}
\vspace{-0.1cm}

We aggregate the diagnosis categories into $k$ clusters by using the spectral state clustering method in Section \ref{method}, for $k = 2,\ldots, 9$. The partition map is further used to build a kernel function \eqref{kernel} for the state space $\mathcal{S}$.
We view $k$, the rank parameter used in state compression, as a measure of the model complexity. It is a tuning parameter that controls the tradeoff between model reduction error and overfitting error.

We conduct both in-sample and out-of-sample experiments. 
In in-sample experiments, we train the treatment policy to minimize cost over the entire data set, and test its performance on the same raw data. In out-of-sample experiments, we use 2-fold cross validation. We split the raw data set into two equal-size segments randomly, and compute the optimal policy using one segment and test its performance on the another segment. Each experiment is repeated randomly for 200,000 times and for all values of $k$. The results are given in Figure \ref{normal}.
In particular, Figure \ref{normal} (a) shows the average cost per episode obtained from in-sample and out-of-sample experiments across values of $k$, as well as their confidence intervels. 

Note that according to \citep{japsen_2016}, 
hospitals have to keep their costs below \$25,565 or they will face a financial charge. We define the episodic excessive cost premium as 
\vspace{-0.1cm}
\begin{align*}
\text{episodic premium}  = \max\{0, \text{episode cost} - 25565\}.
\end{align*}
In Figure \ref{normal} (b) we compare the episodic premium from the in-sample and out-of-sample experiments.

The results of Figure \ref{normal} are consistent with typical in-sample vs out-of-sample error curves for training machine learning models. It is expected that the in-sample error decreases monotonically as the model complexity $k$ increases, however, over-fitting would occur when $k$ is too large. The out-of-sample error curves show that over-fitting occurs when $k>3$, therefore we pick $k=3$ to be the best model complexity for this data set. We take the out-of-sample performance at $k=3$ as the predicted performance of the optimized policy. In this case, the average cost per episode reduces by 7 percent and the excessive cost premium per episode reduces by 33 percent.

\vspace{-0.1cm}
\subsection{Cost Distribution and Tail Improvement}\label{cost}
\vspace{-0.1cm}
 
In Figure \ref{episode cost} we compare the histograms of episodic cost over the raw data and after optimization. The after-optimization cost distribution is obtained by simulation. The percentages of episodes costing over \$30k are reduced significantly. Such a reduction mitigates the financial risk faced by healthcare providers, as they are penalized for costs over \$25,565 according to CMS. The tail improvement is significant for reasons beyond financial risk. It is well known that excessive medical cost in medical treatment is highly correlated to unsuccessful cases, for example post-surgery complications often lead to extra surgeries, higher costs and traumatic discomforts. Therefore the tail cost reduction shown in Figure \ref{episode cost} suggests that the optimized clinical policy may reduce unnecessary medical complications and extra inpatient surgeries. The result implies that the proposed pipeline would not only reduce financial costs but also improve the quality of care.

\begin{figure*}
\centering
\includegraphics[width= 3.8in, height = 1.6in  ]{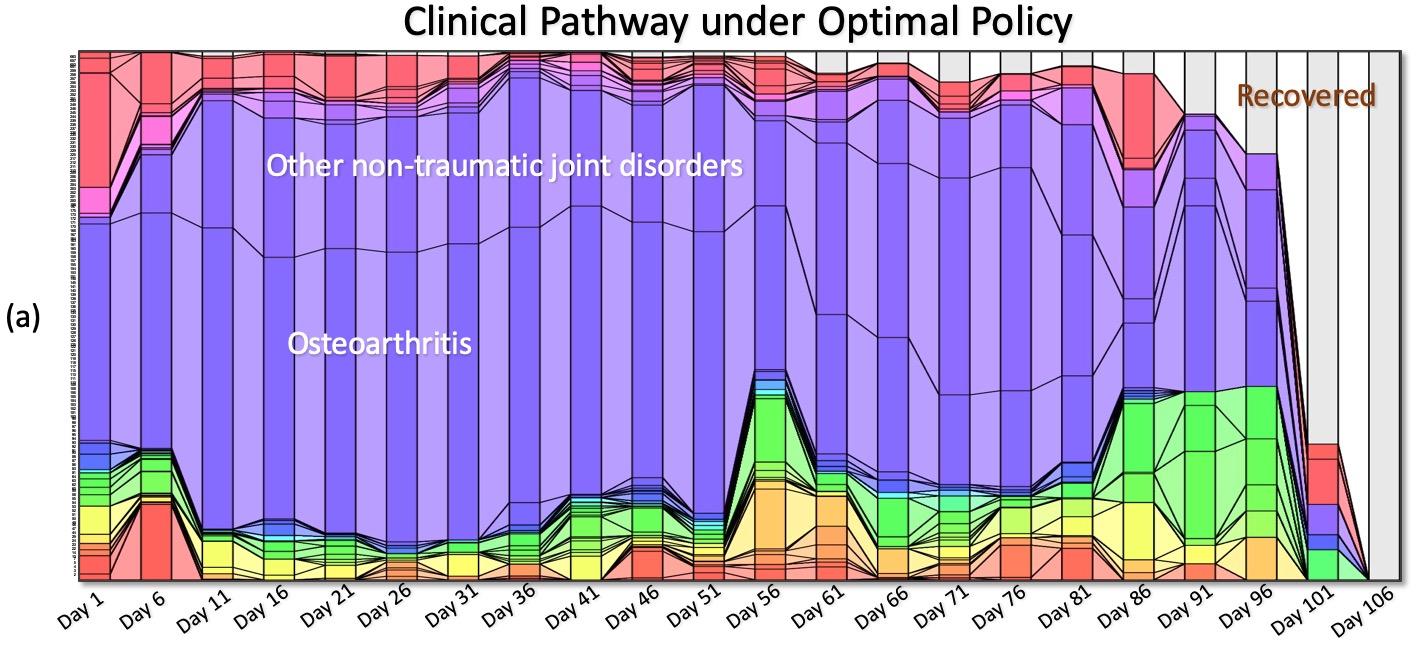}\\
\includegraphics[width= 1.8in, height = 1.15in]{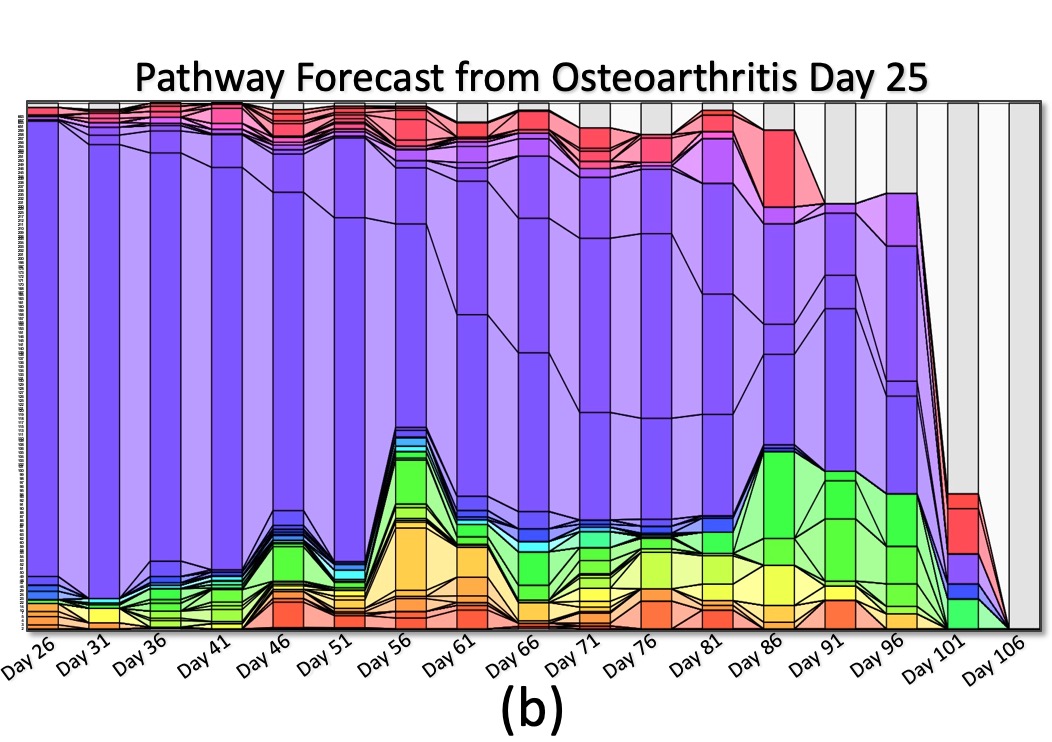}
\includegraphics[width= 1.37in, height = 1.15in]{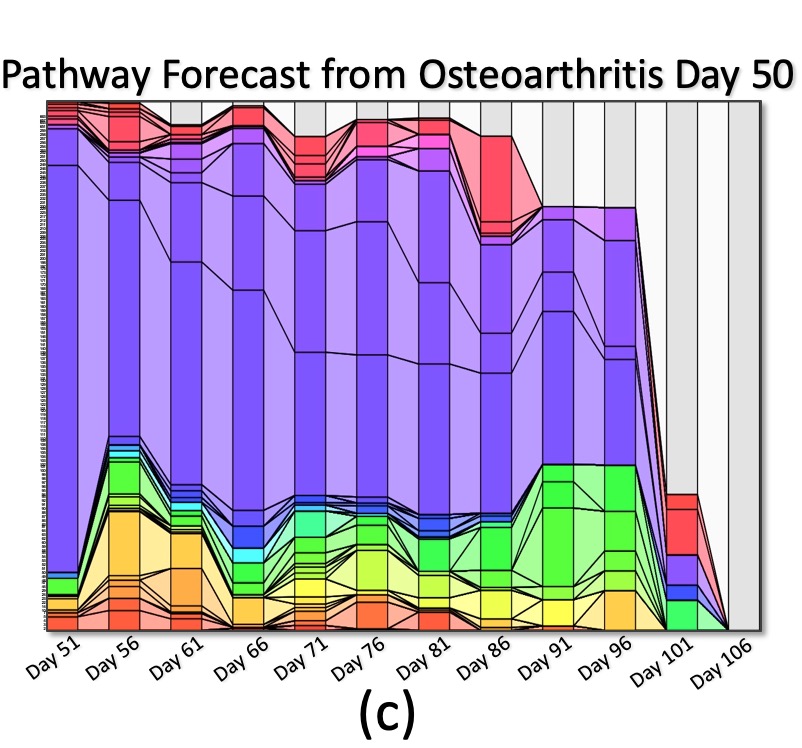}
\includegraphics[width= 0.78in, height = 1.15in]{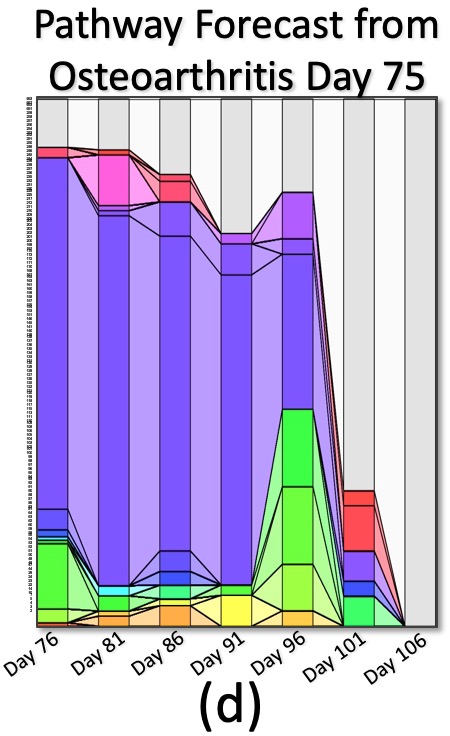}
\includegraphics[width= 1.37in, height = 1.15in]{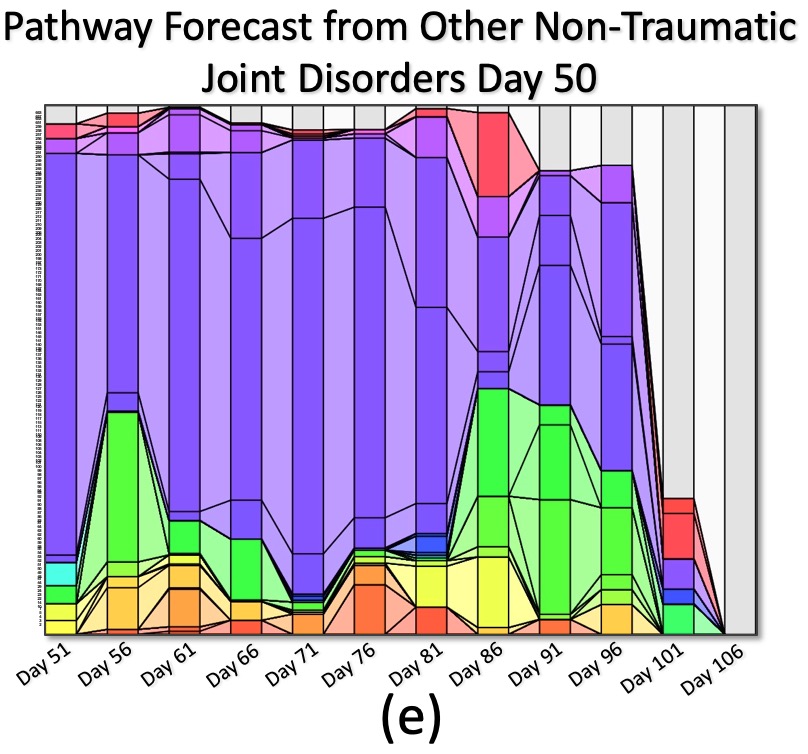}\\
\setlength{\belowcaptionskip}{-10pt}
\caption{\footnotesize{{\bf Clinical pathway forecasts under the optimized treatment policy.} X-axis corresponds to the day of treatment, and Y-axis corresponds to 138 diagnosis categories (each has a unique color). For each day, the bar plot shows the probability distribution of a patient's state over the 138 diagnosis categories. Figure (a) gives the full clinical pathway for a typical patient. Figure (b)(c)(d) give forecasts of the future pathway from day 25, day 50, day 75 respectively, conditioned on the event that the current diagnosis belongs to the osteoarthritis category. Figure (e) illustrated the forecasted pathway from day 50, conditioned on the event that the current diagnosis belongs to the category of non-traumatic joint disorders. }}
\label{pathway}
\end{figure*}

\vspace{-0.2cm}
\subsection{Clinical Pathway Forecast Using The Optimal Policy}\label{clinical}
\vspace{-0.1cm}

Now we have estimated the best clinical policy from data. Our model allows us to predict the entire clinical pathway of a typical knee replacement treatment. We simulate the knee replacement process under the optimized policy based on existing claim episodes. This generates 10,000 sample trajectories. Then based on the sample trajectories, we compute the empirical distribution of diagnosis categories per each day.
 Figure \ref{pathway} (a) illustrates the distribution over states as time progresses, when the optimal policy is used. 
 
Our model can also forecast the future clinical pathway of a patient from any given time and given state during the knee replacement episode. Figure \ref{pathway} (b)(c)(d) illustrates the forecasted pathway from day 25, day 50, day 75,  respectively, conditioned on the event that the current diagnosis belongs to the category of  osteoarthritis. Figure \ref{pathway} (e) illustrated the forecasted pathway from day 50, conditioned on the event that the current diagnosis belongs to the category of non-traumatic joint disorders. We see that the model is able to predict the conditional distributions of future diagnosis and treatments given any particular time and state of the treatment. The forecasts would be useful in terms of making physicians better informed before giving prescriptions.

\vspace{-0.3cm}
\section{Summary}
\vspace{-0.2cm}
This paper provides a reinforcement learning-based pipeline to optimize the knee replacement treatment from medical claims.
It produces a clinical policy via intelligence crowdsourcing from experts and unsupervised state compression. Empirical validation suggests the optimized policy reduces the average cost per episode by \$1,000, which implies a potential savings of $700$ millions each year in the US. We believe this result gives a proof-of-concept that reinforcement learning can significantly improve clinical decisions.
For future work, we will focus on personalizing the knee replacement treatment by using data sets of large scales and higher-resolution models.


\bibliography{neurips_2019_new}
\bibliographystyle{apalike}

\newpage
 
\appendix

\section{Data}\label{appendix}

We use a data set of knee replacement claims from Cedar Gate Technologies. It is a healthcare analytics company that provides prescriptive tools to help healthcare organizations and insurance companies manage their costs and risks.

The data set contains claims records from 37 physicians and 205 unique beneficiaries (patients). It contains 212 episodes, 5946 claims and 9254 entries. Each episode is a full claims record for a patient' pathway from onset to rehabilitation and recovery. Each episode involves one physician, one beneficiary, a total cost, an episode start date, an episode end date, and a sequence of claims.

\textbf{Attributes:} Each entry contains the following attributes:
episode id, beneficiary id, episode start date, episode end
date, episode total cost, physician id, claim id, claim start
date, claim end date, claim cost, procedure code, procedure
category, diagnosis code, diagnosis category.

\textbf{Date:} Each episode has one ``episode start date" and one ``episode end date", and each claim has one ``claim start date" and one ``claim end date". Most claims start and end on the same day. But still, there exists some claims whose ``claim start date" and ``claim end date" are different. 

\textbf{Claim:} Each claims consists of diagnoses, procedures, a start date, an end date and a cost of the claim. There are 597 different procedures, 81 procedure categories, 542 different diagnoses, and 138 diagnosis categories in this data set. Almost all claims involves only one diagnosis and/or only one procedure. For example, see line 6 in Figure \ref{data}, where ``NA" means the corresponding procedure or diagnosis does not exist. {Some claims have more than one entries.}

\textbf{Beneficiaries and Physicians:} There are 37 physicians and 205 beneficiaries in this data set. Each beneficiary is attributed to a unique physician. 


\textbf{Duration and Cost:} The duration of episodes are all around 100 days, because knee replacement is a standardized and well-scheduled process. The number of claims per episode ranges from 8 to 56. 

For confidentiality purposes, we are not allowed to disclose additional information about the data set.

\section{Training Details}
In Section \ref{method}, we solve the problem \eqref{partition} by applying the k-means method with 100 random initializations and choosing the one with the least sum of squares. In Section \ref{project}, we do in-sample experiments as following, Step 1: First get the optimal policy using the whole data, and then simulate it 400 times to calculate the average cost. Step 2: Repeat step 1 five hundreds times. The error bar is the 95\% confidence interval of averages in Step 2. In out-of-sample experiments, Step 1: Split the whole data set into two equal-size segments randomly, and compute the optimal policy using one segments. Step 2: Use the optimal policy from step 1 to simulate episode cost on the remaining data 400 times, and get an average cost. Step 3: Repeat Step 1 and Step 2 five hundreds times. The error bar is the 95\% confidence interval of averages in Step 3. In Section \ref{cost}, we choose $k=3$, and run the same experiments as in Section \ref{project}. In Section \ref{clinical}, we simulate 10,000 the knee replacement process under the optimized policy based on existing claim episodes from day 0, day 25, and day 75. Then based on the sample trajectories, we compute the empirical distribution of diagnosis categories per each day.

\end{document}